\documentclass[journal]{IEEEtran}
\usepackage{cite}

\ifCLASSINFOpdf
  \usepackage[pdftex]{graphicx}
\else
  \usepackage[dvips]{graphicx}
\fi
\usepackage[cmex10]{amsmath}
\usepackage{algorithmic}

\usepackage{array}

\ifCLASSOPTIONcompsoc
  \usepackage[caption=false,font=normalsize,labelfont=sf,textfont=sf]{subfig}
\else
  \usepackage[caption=false,font=footnotesize]{subfig}
\fi
\usepackage{dblfloatfix}

\usepackage{booktabs}

\usepackage{url}

\usepackage{float}
\usepackage{multirow}
\usepackage{enumerate}
\usepackage{amsthm}
\usepackage{nidanfloat}
\usepackage{threeparttable}
\newtheoremstyle{mystyle}
  {}
  {}
  {\itshape}
  {}
  {\bfseries}
  {.}
  { }
  {}
\theoremstyle{mystyle}

\newtheorem{remark}{Remark}

\newenvironment{talign}
 {\align}
 {\endalign}
\newenvironment{talign*}
 {\csname align*\endcsname}
 {\endalign}
\usepackage{arydshln}
\usepackage[colorinlistoftodos]{todonotes}
\usepackage[normalem]{ulem}

\begin{document}
%
\title{Class-Incremental Learning for Wireless Device Identification in IoT}



\author{Yongxin~Liu,
        Jian~Wang,
        Jianqiang~Li,
        Shuteng~Niu, and
        Houbing~Song,~\IEEEmembership{Senior Member,~IEEE}
\thanks{Yongxin Liu, Jian Wang, Shuteng Niu, and Houbing Song are with the Security and Optimization for Networked Globe Laboratory (SONG Lab), Embry-Riddle Aeronautical University, Daytona Beach, FL 32114 USA}
\thanks{Jianqiang Li is with the College of Computer Science and Software Engineering, Shenzhen University, Shenzhen, Guangdong 518060 China}
\thanks{Corresponding author: Houbing Song}
\thanks{Manuscript received March 22, 2021; revised XXX. }}

\markboth{IEEE Internet of Things Journal,~Vol.~11, No.~4, May~2021}%
{Shell \MakeLowercase{\textit{et al.}}: Bare Demo of IEEEtran.cls for Journals}
%



\IEEEtitleabstractindextext{%
\begin{abstract}
\textcolor{black}{
Deep Learning (DL) has been utilized pervasively in the Internet of Things (IoT). One typical application of DL in IoT is device identification from wireless signals, namely Non-cryptographic Device Identification (NDI). However, learning components in NDI systems have to evolve to adapt to operational variations, such a paradigm is termed as Incremental Learning (IL). Various IL algorithms have been proposed and many of them require dedicated space to store the increasing amount of historical data, and therefore, they are not suitable for IoT or mobile applications. However, conventional IL schemes can not provide satisfying performance when historical data are not available. In this paper, we address the IL problem in NDI from a new perspective, firstly, we provide a new metric to measure the degree of topological maturity of DNN models from the degree of conflict of class-specific fingerprints. We discover that an important cause for performance degradation in IL enabled NDI is owing to the conflict of devices' fingerprints. Second, we also show that the conventional IL schemes can lead to low topological maturity of DNN models in NDI systems. Thirdly, we propose a new Channel Separation Enabled Incremental Learning (CSIL) scheme without using historical data, in which our strategy can automatically separate devices' fingerprints in different learning stages and avoid potential conflict. Finally, We evaluated the effectiveness of the proposed framework using real data from ADS-B (Automatic Dependent Surveillance-Broadcast), an application of IoT in aviation. The proposed framework has the potential to be applied to accurate identification of IoT devices in a variety of IoT applications and services. Data and code available at IEEE Dataport (DOI: 10.21227/1bxc-ke87) and \url{https://github.com/pcwhy/CSIL}}.
\end{abstract}

}

\maketitle

\IEEEdisplaynontitleabstractindextext

%
\IEEEpeerreviewmaketitle

\section{Introduction}
%
%
%
%
\textcolor{black}{
The Internet of Things (IoT) is characterized by the interconnection and interaction of smart objects (objects or devices with embedded sensors, onboard data processing capabilities, and means of communication) to provide applications and services that would otherwise not be possible \cite{IIOT17,7406686,taheri2020fed}. The convergence of sensors, actuators, information, and communication technologies in IoT produces massive amounts of data that need to be sifted through to facilitate reasonably accurate decision-making and control \cite{BDA19, song2016cyber,song2017smart,chu2020intelligent}. A typical way to implement smart decision functionality in IoT is by integrating learning-enabled components through Deep Learning (DL) and Deep Neural Networks (DNNs). One typical application of DNNs in IoT is the passive identification of IoT devices through their wireless signals for Non-cryptographic Device Identification (NDI) and Physical Layer authentication \cite{liu2021machine,song2017security,8897627}. DL and DNNs are effective in wireless device identification under various scenarios, however, DNN models in these applications need to continuous evolving to adapt to operational variations as new devices (as new classes) are emerging. Such a continuous evolving scheme is termed as Lifelong or Incremental Learning (IL).
}
\textcolor{black}{
Conventional approaches require periodic retraining to update DNNs. In this paradigm, DNNs are initialized from scratch and trained with all past and present devices' signals. Even though the best accuracies are guaranteed in these Non-Incremental Learning (Non-IL) schemes, the memory consumption and training time can grow drastically as new devices are added in. Therefore, there is a need for IL with a reasonable balance between accuracy, memory consumption, and training efficiency. In IoT, less or zero memory for historical data are preferred during the continuous evolving \cite{belouadah2020initial}.
}

\textcolor{black}{
Compared to conventional non-incremental learning (non-IL) schemes, DNN models can only use a very small proportion or even none of the data from the previous stages, a.k.a. old tasks, while they are trained to recognize new devices. The absence of data from old tasks results in \textit{Catastrophic Forgetting}, a phenomenon of significant degradation of accuracy after training on new tasks. IL has become an emerging topic in machine learning, however, many of the methods are not adaptable in IoT. For example, some works require storing specifically chosen old data, and can consume a large amount of memory \cite{liu2020mnemonics} gradually. Other works require incrementally training task-related generative models for knowledge replay, but these generative models require notorious efforts \cite{liu2020generative}. In addition, there are several attempts to either use regularization or knowledge distillation to implement memoryless methods to prevent DNNs from forgetting \cite{belouadah2020initial}. Balancing between learning and forgetting is difficult, especially when the internal mechanism of catastrophic forgetting is not yet clear. Besides, there is a lack of theoretic explanation to explore the difference between the key characteristics between IL and regularly trained models.}

\textcolor{black}{
In this paper, we explored the topological properties of fingerprints in the final classification layers of DNN-enabled wireless device identification models after IL and regular training, we discovered that the main cause of catastrophic forgetting is due to the nonoptimal distribution of feature vectors and their reprentatives (fingerprints) in the latent space. Based on the discoveries, we designed an enhanced IL scheme, the Channel Separation Enabled Incremental Learning (CSIL), for wireless device identification systems. We manually introduced separations in representative spaces between different tasks (learning stages). The effectiveness of the proposed framework in massive signal recognition and improving the incremental learning performance has been demonstrated. The contributions of this paper are as follows:}
\begin{itemize}
    \item \textcolor{black}{We provide a new metric, the Degree of Conflict (DoC), to quantitatively analyze the topological maturity of DNN models. Using this metric, we discover that DNN models trained by conventional IL mechanisms are with low topological maturity. This metric is helpful in understanding the internal mechanisms of DNNs. }
    \item We provide a new perspective for the causality, the conflict of fingerprints, to explain the catastrophic forgetting in DNN models.
    \item We provide an enhanced IL strategy, CSIL for incremental learning for DNN-enabled IoT device identification systems and test the CSIL mechanism using real signal datasets.
    
\end{itemize}
Our research offers not only a solution for accurate identification of IoT devices, but also useful for future development of IL for DNNs. To our best knowledge, this is the first study that jointly explores DNN and IL in Signal Intelligence Applications. Right before the publication of this work, we realized that our algorithm actually has solid evidence from the most recent advancement of neural science \cite{libby2021rotational}. We share some similar findings as in \cite{libby2021rotational}, but from a totally different perspective and a non-biological road map. In addition, we provide the mathematical proof and are delighted to find an elegant connection between biological and artificial intelligence.

The remainder of this paper is organized as follows: A literature review of wireless device identification and incremental learning is presented in Section~\ref{sectRW}. We formulate our problem in Section~\ref{sectPD} with the methodology presented in Section~\ref{sectMM}. Performance evaluation is presented in Section~\ref{sectEED} with conclusions in Section~\ref{sectCC}.

\section{Related works}
\label{sectRW}
In this section, we will provide a brief review of wireless device identification in IoT and Incremental Learning in deep learning.
\subsection{Wireless device identification in IoT}

Specific device identification is emerging as a solution to Physical layer security of IoT. The methods aim to recognize IoT devices based solely on their signals. Corresponding methods can be classified into two categories: specific feature based and deep learning based approaches. 

The specific feature based approaches require human efforts to discover distinctive features for device identification. The methods rely on the fact that there are various manufacturing imperfectnesses in wireless devices' RF frontends. These imperfectnesses do not degrade the communication quality but can be exploited to identify each transmitter uniquely. Those features are named Physical Unclonable Features (PUF) \cite{chatterjee2018rf,herder2014physical}). Some works assume that the statistical properties of noise or errors could uniquely profile wireless devices. In \cite{azarmehr2017wireless}, the authors show that the phase error of Phase Lock Loop in transmitters can provide promising results even with low Signal-to-Noise Ratio (SNR). In \cite{zhuang2018fbsleuth}, the authors use the error between received signals and theoretical templates and use time-frequency features to fingerprint different transmitters. In \cite{peng2019deep}, the authors employ the differential constellation trace figure (DCTF) to capture the time-varying modulation error of Zigbee devices. They then develop their low-overhead classifier to identify 54 Zigbee devices. 


Feature-based approaches require efforts to manually extract features or high-order statistics for different scenarios. Therefore, more effortless and versatile methods are required. Deep Neural Networks (DNNs) are frequently used as a general-purpose blackbox for pattern recognition. and can significantly reduce the hardship of manual feature discovery. In \cite{yu2019radio}, the authors provide a novel method that performs signal denoising and emitter identification simultaneously using an autoencoder and a Convolution Neural Network (CNN). Their solution shows promising results even with low SNR. Similar work in \cite{huang2017communication} employs a stacked denoising auto-encoder and shows similar results. DNNs perform well even on raw signals. In \cite{riyaz2018deep}, the authors provide an optimized Deep Convolutional Neural Network to classify SDR-based emitters in 802.11AC channels, they show that, even by using raw signals without feature engineering, CNN surpasses the best performance of conventional statistical learning methods. In \cite{morin2019transmitter}, neural networks were trained on raw IQ samples using the open dataset from CorteXlab. Their works also show similar results. Compared with specific feature based approaches, deep neural networks dramatically reduce the requirement of domain knowledge and the quality of fingerprints. 
\subsection{Incremental Learning in Deep Neural Networks}
In general, DNNs are effective in non-cryptographic wireless device identification. However, a DL enabled wireless device identifier has to learn new devices' characteristics during its life cycle. Such functionalities are defined as lifelong learning or Incremental Learning (IL). 

Conventionally, Transfer Learning (TL) are applied, neural networks are pretrained in the lab and then fine-tuned for deployment using specific data \cite{niu2020decade, niu2020transfer}. In TL, the learning components can forget a large proportion of the knowledge they learn in the lab and adapt to new scenarios. In Incremental Learning (IL), neural networks are trained incrementally as new data come in progressively \cite{parisi2019continual}. CL does not allow neural networks to forget what they have learned in the early stages compared with TL. Therefore, TL is useful when deploying new systems, and CL is useful in regular software updates and maintenance. The strategies to implement CL for DNN are in three folds:

\textit{Knowledge replay:} An intuitive solution for CL is to replay data from old tasks while training neural networks for new tasks. However, such a solution requires long training time and large memory consumption. Besides, one can hardly judge how many old samples are enough to catch sufficient variations. Therefore, some studies employ generative networks or exemplars to replay data from old tasks \cite{shin2017continual}. In \cite{shin2017continual}, Generative Adversarial Network (GAN) based scholar networks are proposed to generate old samples and mixed with the current task. In this way, the deep neural network could be trained on various data without using huge memories to retain old training data\cite{van2018generative}. However, data generators are not easy to train and retaining old data will gradually consume a lot of memory and thus not yet a good choice for wireless device identification system in IoT.

\textit{Regularization: }Initially, regularization is employed to prevent models from overfitting by penalizing the magnitude of parameters \cite{girosi1995regularization}. In CL, regularization is employed to prevent models from changing dramatically. In this way, the knowledge (represented by weights) learned from the old tasks will be less likely to vanish when trained on new tasks. In Elastic Weight Consolidation (EWC) \cite{kirkpatrick2017overcoming}, the algorithms identify important connections and protect them from changing dramatically, in which noncritical connections are used to learn new tasks. Regularization does not require storing old samples or data generators but may not have a high accuracy as knowledge replay.

\textit{Dynamic network expansion: }Network expansion strategies lock the weights of existing connections and supplement additional structures for new tasks. For instance, the Dynamic Expanding Network (DEN) \cite{yoon2017lifelong} algorithm first trains an existing network on a new dataset with regularization. The algorithm compares the weights of each neuron to identify task-relevant units. Finally, critical neurons are duplicated and to allow network capacity expansion progressive. The problem for the method is the need to know the task information to select appropriate data flow paths.

Incremental learning, especially memoryless class incremental, is rarely covered in signal intelligence systems, such as wireless device identification, thus motivating our research.

\section{Methodology}
\label{sectPD}
\subsection{Zero-bias Deep Neural Network for Wireless Device Identification}
\label{sectZbDNN}
We focus on deriving a protocol-agnostic wireless device identification system with incremental learning capability. Suppose that the radio signal from a specific device $i$, is denoted as $\hat{m}_i(t)$:
\begin{align}
\hat{m}_i(t) = m_i(t) + \delta_i(t) = I(t)+ \boldsymbol{i} \cdot Q(t)
\end{align}
where $m_i(t)$ is the message while the residual, $\delta_i(t)$, is exploited to recognize a wireless device. $\delta_i(t)$ is also defined as the pseudo noise signal. If $\delta_i(t)$ is uncorrelated with messages $m_i(t)$, the recognition algorithm can be protocol-agnostic. In this work, we use a Software-Defined Radio (SDR) receiver (USRP B210) for signal reception, therefore, $I(t)$ and $Q(t)$ are the in-phase and quadrature components respectively. 

Suppose $m_i(t)$ is successfully extracted from $\hat{m}_i(t)$ and we also extract the frequency domain features from the pseudo noise as:
\begin{align}
    \delta_i(\omega) = FFT[\hat{m}_{i}(t)]-FFT[m_{i}(t)]
\end{align}{}
where $r_{i}(t)$ is the reconstructed rational baseband signal. Please be noted that $\hat{m}_{i}(t)$ is complex-valued (QPSK) while $r_{i}(t)$ can be real-valued (2FSK, 2PSK, and etc.). %

\begin{figure}[h]
\centering
\includegraphics[width=0.85\linewidth]{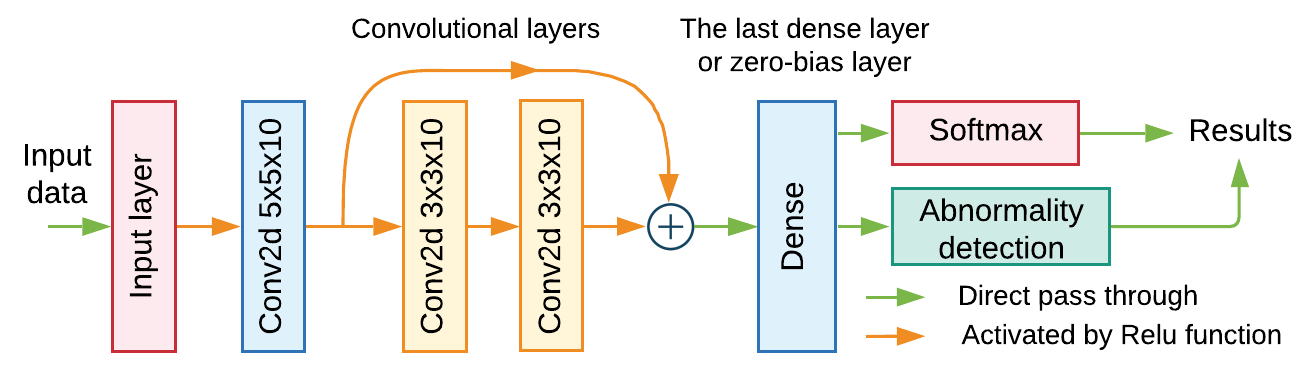}
\caption{Deep neural network for wireless device identification. }
\label{figLearningArch}
\end{figure}

We convert $\delta_i(\omega)$ into a magnitude sequence ($||\delta_j(\omega)||$), namely, Mag.-Freq. residuals, and a phase sequence ($\angle \delta_i(\omega)$), namely Phase-Freq. residuals, respectively. These three types of signal features are passed through a Deep Neural Network based wireless device identification model, depicted in Figure~\ref{figLearningArch}.

\begin{figure}[]
\centering
\includegraphics[width=0.85\linewidth]{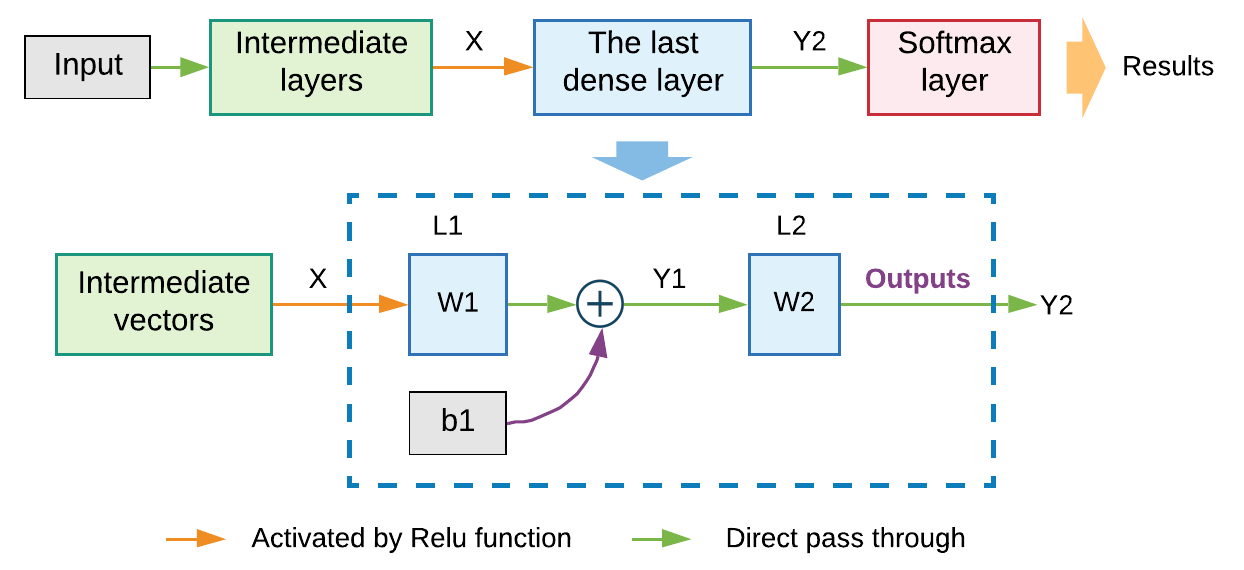}
\caption{Data flow of zero-bias deep neural networks.}
\label{figZerobiasDense}
\end{figure}
We have discovered that the last dense layer of a DNN classifier performs the nearest neighbor matching with biases and preferabilities using cosine similarity, We also show that a DNN classifier's accuracy will not be impaired if we replace its last dense layer with a zero-bias dense layer \cite{liu2020zero}, in which the decision biases and preferabilities are eliminated. We can denote its mechanism as (also in Figure~\ref{figZerobiasDense}):
\textcolor{black}{
\begin{align}
\label{eqZeroBiasDenseLayer}
    \notag\boldsymbol{Y_1}(\boldsymbol{X}) &= \boldsymbol{W_0}\boldsymbol{X} + \boldsymbol{b}\\
    \boldsymbol{Y_2}(\boldsymbol{X}) &=  cosDistance(\boldsymbol{Y_1}, \boldsymbol{W_1})
\end{align}
}
\textcolor{black}{
Where $\boldsymbol{X}$ is the output of the prior convolution layers, a.k.a., feature vectors. $\boldsymbol{X}$ is an $N_0$ by $q$ matrix, where $N_0$ denotes the number of features while $q$ denotes the batch size. $\boldsymbol{W}_0$ is an $N_1$ by $N_0$ matrix where $N_1$ denotes the number of embedded features. $\boldsymbol{W_1}$ is a matrix to store fingerprints of different classes, namely the similarity matching layer and it is a $C$ by $N_1$ matrix in which $C$ denotes the number of classes, we set $N_1 = 2C$ in this paper. Please be noted that in $\boldsymbol{W_1}$, each row represents a fingerprint of corresponding class whilst in $\boldsymbol{Y}_1$ each column represents a feature vector in the latent space. Intuitively, the last dense layer is spitted into two layers, $L_1$ for feature embedding and $L_2$ for similarity matching. The cosine similarity matching is denoted as:
}
\textcolor{black}{
\begin{align}
\label{eqCosine}
    cosDistance(\boldsymbol{Y_1}, \boldsymbol{W_1}) =\boldsymbol{RU(W_1)} \times \boldsymbol{CU(Y_1)}
\end{align} 
}
\textcolor{black}{
Where $\boldsymbol{RU(\cdot)}$ and $\boldsymbol{CU(\cdot)}$ denote deriving column-wise and row-wise direction vectors (vectors' magnitudes are normalized to one) of their inputs. Our prior results \cite{liu2020zero, liu2020deep} prove that the zero-bias dense layer can work seamlessly with backpropagation mechanisms and trained using regular loss functions (e.g., binary crossentropy, etc.). Please be noted that even if $L_2$ can be replaced by a regular dense layer, it can also be viewed as a similarity matching layer, but the matching results are weighted and biased \cite{liu2020zero}.}
\subsection{Optimal separation of fingerprints}

Intuitively, if the devices' fingerprints are distantly separated in the latent space, we will have less chance to confuse them. To quantify the separation, the sum of the mutual cosine distances of all devices' fingerprints in a classification model can be defined as:
\begin{align}
\label{eqSumDev}
    \notag \textstyle TD&(\boldsymbol{f_1},\cdots,\boldsymbol{f_C}) = \sum^{C}_{i=1, j < i} CosineDistance(\boldsymbol{f_i},\boldsymbol{f_j})\\
    &= \sum^{C}_{i=1, j < i} x^{(1)}_{i}x^{(1)}_{j}+x^{(2)}_{i}x^{(2)}_{j}+\cdots+x^{(N1)}_{i}x^{(N1)}_{j}
\end{align}
where $\boldsymbol{f_i} = (x^{(1)}_{i},x^{(2)}_{i},\cdots,x^{(N1)}_{i})$ and $\boldsymbol{f_j} = (x^{(1)}_{j},x^{(2)}_{j},\cdots,x^{(N1)}_{j})$ are devices' fingerprint vectors. Suppose we have $C$ devices with $N_1$-D fingerprint vectors. Noted that the fingerprints have been normalized into unit vectors. Therefore, if we need to find the optimal value of $TD(\cdot)$, we need to incorporate the constraints: 
\begin{align}
\label{eqUnitFP}
\textstyle \forall i,~ g(\boldsymbol{f_i})=\sum^{N1}_{d=1}(x^{(d)}_i)^2-1=0
\end{align}
Equation \ref{eqSumDev} has now become a constrained optimization problem. We solve this constrained optimization problem with the Lagrange Multiplier as:
\begin{talign}
\notag L(\boldsymbol{f_1},&\cdots,\boldsymbol{f_C},\lambda_1,\cdots,\lambda_C) \\
&= TD(\boldsymbol{f_1},\cdots,\boldsymbol{f_C}) - \sum^C_{i=1}\lambda_i g(\boldsymbol{f}_i)
\end{talign}
And we need to solve:
\textcolor{black}{
\begin{talign}
    \underset{x^{(1)}_1\cdots x^{(N1)}_1,\cdots,x^{1}_C \cdots x^{N1}_C,\lambda_1 \cdots \lambda_i }{\nabla}	L( \boldsymbol{f_1} \cdots \boldsymbol{f_C},\lambda_1 \cdots \lambda_C)=0
\end{talign}
}
Which results in a linear system of equations. For each $k$th ($k=1\cdots N_1$) dimension of fingerprint vectors $x^{(k)}_1,\cdots,x^{(k)}_C$, we have:
\begin{talign}
\label{eqSolveLbd}
\notag\frac{\partial L}{x^{(k)}_1} =-2\lambda_1 x_1^{(k)} &+ \sum^C_{i=1,i\neq 1}x^{(i)}_1 = 0\\
\vdots\text{~~~~~~~~~~~~~~~~~}&\cdots\text{~~~~~~~~~~~~~~~~~}\vdots\\
\notag\frac{\partial L}{x^{(k)}_{C}} =-2\lambda_C x^{(k)}_C &+ \sum^C_{i=1,i\neq C}x^{(i)}_C = 0
\end{talign}
This is a homogeneous system of equations, and it is unlikely that it only has a trivial solution (zeros). Hence, $\lambda_1=\lambda_2=\cdots=\lambda_C=-0.5$ and \textcolor{black}{Equation \ref{eqSolveLbd} can be converted into one equation:}
\begin{talign}
\label{eqFingerprintEq}
\sum^{C}_{i=1}x^{(k)}_i = 0
\end{talign}
We square Equation \ref{eqFingerprintEq} and expand it. According to Multinomial Theorem \cite{Multinom81:online} we have:
\begin{talign}
\label{eqFingerprintEqSquared}
\sum^{C}_{i=1}(x^{(k)}_i)^2 + 2\sum^{C}_{ n= 1, m <n }x^{(k)}_n x^{(k)}_m = 0
\end{talign}
Given that $k=1\cdots N_1$, we have $N_1$ Equations with an identical form of Equation \ref{eqFingerprintEqSquared}. By summing them up, we have:
\begin{align}
\label{eqFingerprintEqSquared2}
\sum^{N1}_{k=1}\sum^{C}_{i=1}(x^{(k)}_i)^2 + 2\sum^{N1}_{k=1}\sum^{C}_{ n= 1, m <n }x^{(k)}_n x^{(k)}_m = 0
\end{align}
On the left of Equation \ref{eqFingerprintEqSquared2}, the first part is the sum of the magnitude of fingerprint vectors. And its value is $C$. The second part is exactly two times $TD(\boldsymbol{f_1},\cdots,\boldsymbol{f_C})$ in Equation \ref{eqSumDev}. Therefore, we have:
\begin{remark}
\label{rmSumDevResult}
The sum of the mutual cosine distances of classes' fingerprints of the zero-bias DNN at a converging point is a predictable constant:
\begin{align}
\label{eqSumDevResult}
TD&(\boldsymbol{f_1},\cdots,\boldsymbol{f_C})=-\frac{C}{2}    
\end{align}
\end{remark}
When such a value is reached, the separation of fingerprints are maximized in the latent space, indicating the lowest degree of conflict. \textcolor{black}{We will use the term \textit{Degree of Conflict (DoC)} to describe the characteristic of the zero-bias DNN. Noted that the range of DoC is from $-\frac{C}{2}$ to $\frac{C(C-1)}{2}$. The maximum value is reached when all fingerprints collide into one single vector.}

\begin{figure}[]
\centering  
\subfloat[]
{%
    \includegraphics[width=0.52\linewidth]{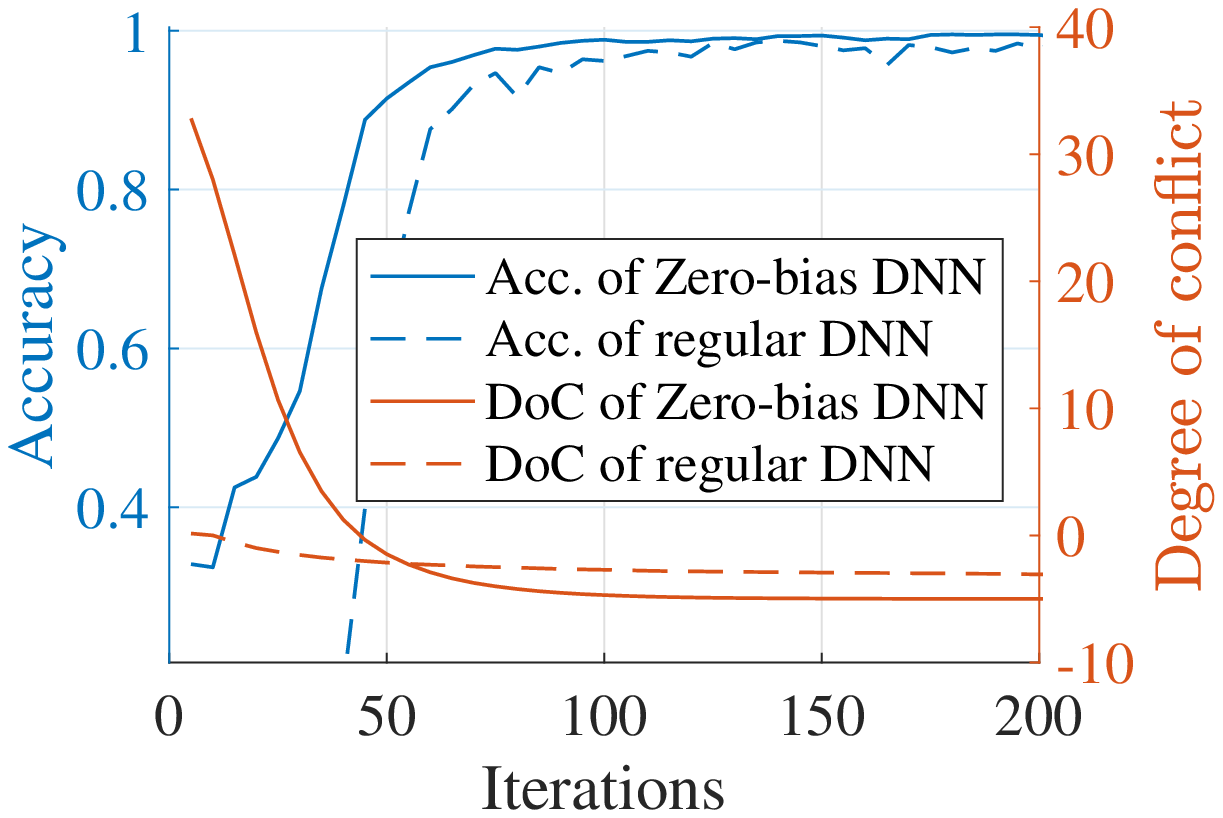}
    \label{figTrainVsSumDev}
}
\subfloat[]
{%
    \includegraphics[width=0.52\linewidth]{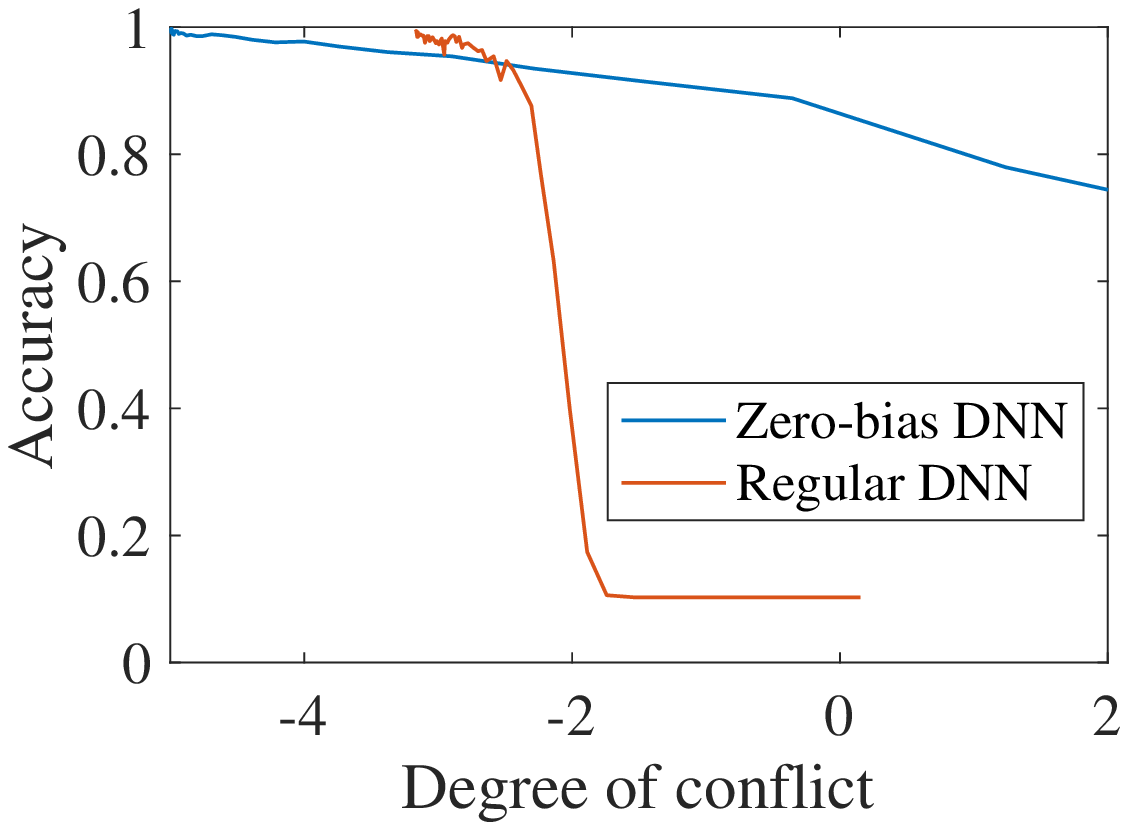}
    \label{figAccVsSumDev}
}
\caption{A comparison of regular and zero-bias DNN considering degree of conflicts and training accuracy.}
\label{figDoCZbFC}
\end{figure}

To demonstrate the Remark \ref{rmSumDevResult}, we use a simple DNN \cite{matlabMinst} with two configurations. In the first configuration, a regular dense layer is applied for the final classification. And in the second configuration, the last dense layer is modified to perform the cosine similarity matching as in Equation \ref{eqZeroBiasDenseLayer}. The two models are trained on the hand-written digit dataset (MNIST). And the change of DoC and accuracy during training are depicted in Figure~\ref{figDoCZbFC}. In Figure~\ref{figTrainVsSumDev}, the degree of conflict of zero-bias DNN model converges to the predicted optimal constant $-\frac{10}{2}=-5$. However, in the regular DNN model, the metric stops at a nonoptimal point, $-3$. Notably, higher accuracy could sometimes reflect a lower DoC between fingerprints. \textcolor{black}{Figure~\ref{figAccVsSumDev} also reveals that the zero-bias DNN model is less sensitive to the variation of DoC.}

\subsection{Analyzing the catastrophic forgetting from the conflict of fingerprints}
\label{sectConflictsIL}
With the cosine similarity matching mechanism. One may assume that incremental learning can be performed by simply inserting new fingerprints. However, we discover that such an intuitive method could cause significant performance degradation. An important factor to cause the performance degradation is the conflict of fingerprints.

\begin{figure}[]
    \centering
    \includegraphics[width =\linewidth]{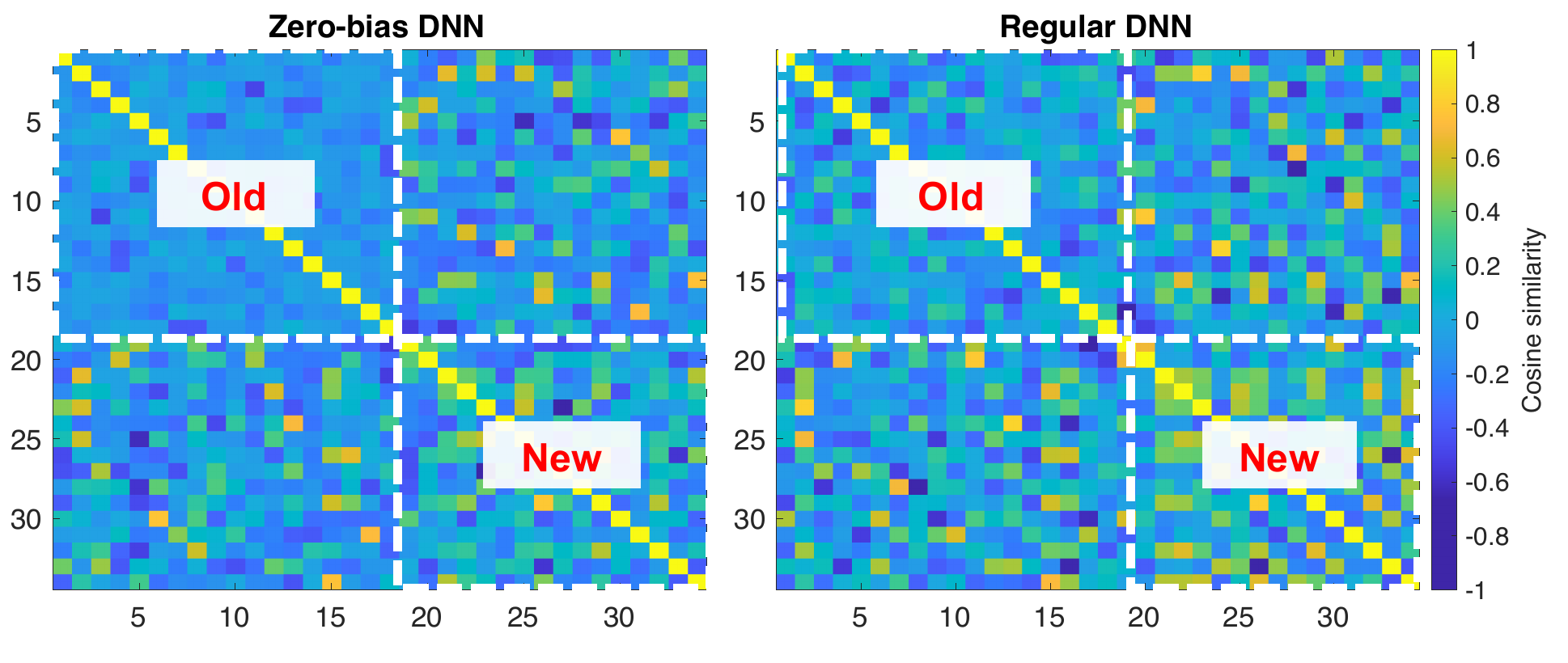}
    \caption{Distance matrix of fingerprints after inserting new fingerprints and finetuning.}
    \label{figCosineSimAftIL}
\end{figure}

To exemplify this phenomenon, we use two DNN models with an architecture specified in Figure~\ref{figLearningArch}, we modify their last dense layers as in Figure~\ref{figZerobiasDense}, we use cosine similarity matching in $L_2$ for the first DNN model and use regular dense layer for $L_2$ for the second one, and therefore, the second DNN is a regular DNN. \textcolor{black}{The two models are tested using a two-stage incremental learning scheme: a) in the first learning stage, the two models are first trained on a wireless signal identification dataset \cite{gt9v-kz32-20} to classify 18 most frequently seen wireless devices. b) Before the second learning stage, we insert the hypothetic fingerprints (generated by averaging feature vectors) of the remaining 16 new devices into their similarity matching layers and freeze all prior layers and fingerprints of learned devices. c) In the second stage, the IL stage, we finetune the newly inserted fingerprints. After the two-stage learning, the cosine similarity matrix of fingerprints in the two models before and after incremental learning is compared in Figure~\ref{figCosineSimAftIL}. }

\textcolor{black}{
The results in Figure~\ref{figCosineSimAftIL} indicate a typical conflict scheme. On the one hand, some fingerprints of the newly learned classes (devices) are less distantly separated as they have higher cosine similarities. On the other hand, some new devices' fingerprints have high cosine similarities with old devices' fingerprints. These two factors jointly cause conflict and confusion. A more detail comparison is provided in Table~\ref{tabCompDoM}. The two models degrees of maturity after IL are far from the expected optimal value. And DoC of the new fingerprints are also far from optimal. 
}

\begin{table}[h]
\centering
\caption{A comparison of the degree of maturity of DNN models before and after incremental learning.}
\label{tabCompDoM}
\resizebox{0.49\textwidth}{!}{%
\begin{tabular}{@{}ccccc@{}}
\toprule
\begin{tabular}[c]{@{}c@{}}DNN\\ models\end{tabular} &
  \begin{tabular}[c]{@{}c@{}}DoC (Acc.) \\ initial training\end{tabular} &
  \begin{tabular}[c]{@{}c@{}}DoC (Acc.)\\ a.f.t. finetuning\end{tabular} &
  \begin{tabular}[c]{@{}c@{}}DoC of new\\ fingerprints\end{tabular} &
  \begin{tabular}[c]{@{}c@{}}Acc. on new \\ / old task\end{tabular} \\ \midrule
Regular       & -8.083 (90.54) & -1.16 (65.2) & 9.05 & 75.5 / 54.2 \\
Zero-bias     & -8.96 (92.85)  & -4.3 (84.2)  & 4.03 & 76.2 / 91.3 \\ \midrule
Optimal value & -9             & -18 (92.2)   & -8   & 92.2 / 93.1 \\ \bottomrule
\end{tabular}%
}
\end{table}

Interestingly, the zero-bias DNN outperforms the regular DNN considering less catastrophic forgetting, herein, we will implement our incremental learning algorithm based on zero-bias DNN. Even though zero-bias DNN has an advantage in IL, we claim that:
\begin{remark}
\label{rmAdjustOld}
Without readjustment of old fingerprints or proper separation between old and new fingerprints, the conflict between fingerprints can not be resolved. Under such criteria, the resulting DNN's performance will not be comparable to training with all data from scratch.
\end{remark}
\begin{proof}
Suppose that we have $N_1$ classes at the initial stage and $m$ new classes to learn afterwards. We define that the averaged cosine distance between $N_1$ fingerprints is $\Bar{D}_0$, according to Remark~\ref{rmSumDevResult}, after initial training we have:
\begin{align}
    \frac{N_1(N_1-1)}{2}\overline{D_0}=-\frac{N_1}{2}\text{~and~}\overline{D_0} = -\frac{1}{N_1-1}
\end{align}
When we have $N_1 + m$ classes, $\Bar{D}_0$ has to become:
\begin{align}
    \overline{D_1}=-\frac{1}{N_1+m-1}
\end{align}
It means that if the classes' fingerprints are to be distantly and uniformly separated, the averaged angles of all old fingerprints need to be reduced while learning new classes. This requirement can not be satisfied if the old fingerprints are locked or prevented from changing. When the prior layers are locked, the distribution of feature vectors in the latent space is fixed, simply reducing the separation of fingerprints in old classes will increase the degree of conflict and cause performance degradation, as depicted in Figure~\ref{figAccVsSumDev}. 
\end{proof}

And if $N_1 + m$ gets larger, $\overline{D_1}$ will approximate zero, thus the averaged separation angles between fingerprints should approximate 90 degrees, that is, orthogonal. Therefore, we believe that there will potentially be some improvement if we can properly separate the fingerprints of old and new classes into different topological spaces to avoid conflict.


\subsection{Channel separation enabled incremental learning}
\label{sectMM}
\begin{figure}[h]
    \centering
    \includegraphics[width =0.85\linewidth]{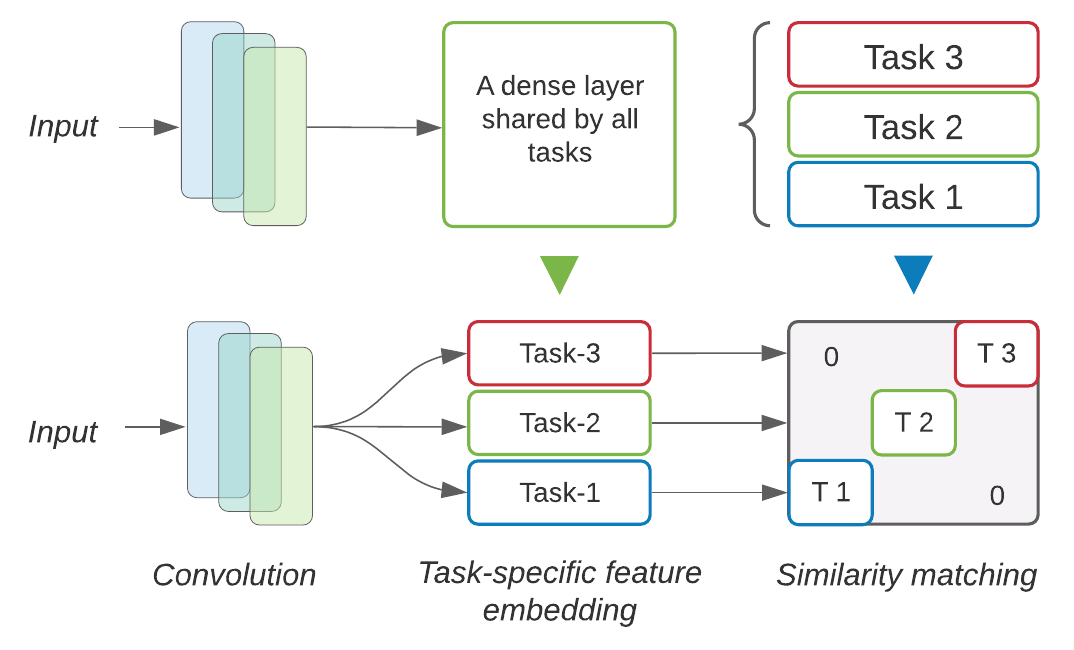}
    \caption{\textcolor{black}{Channel separation for incremental learning}}
    \label{figChannelSeparation}
\end{figure}

To resolve the conflict of fingerprints, we proposed the Channel Separation Enabled Incremental Learning (CSIL), an integral approach incorporating dimension expansion and channel separation as depicted in Figure~\ref{figChannelSeparation}. Intuitively, the merits of this approach are: a) we let the fingerprints learned at different stages to automatically use their task-specific proportions (channels) of parameters in the feature embedding layer. b) We control the directions of fingerprints and force the fingerprints learned from different stages to be orthogonally separated.

\begin{figure}[]
    \centering
    \includegraphics[width =\linewidth]{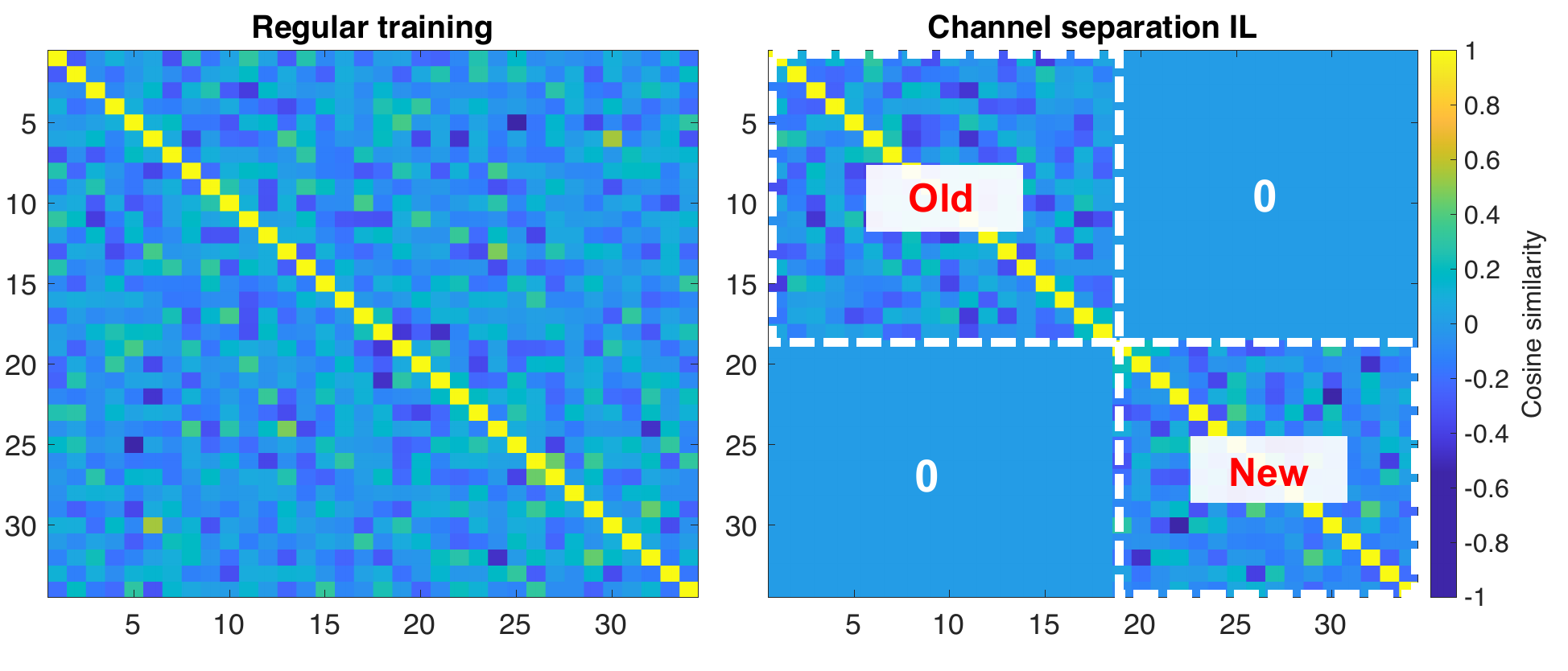}
    \caption{Distance matrix of fingerprints after regular training and CSIL. }
    \label{figCSILSimMatrix}
\end{figure}

At the initial stage, namely \textit{stage-0}, we train a zero-bias DNN as normal. When at the $k$th learning stage, \textit{stage-k}. We first expand the feature embedding layer's weight matrix as:
\begin{talign}
\label{eqILFeatureEmbed}
\boldsymbol{W_0^{(k)}} = 
\left[
    \begin{array}{c;{2pt/2pt}c}
        \boldsymbol{W^{(k-1)}_0} & \boldsymbol{w^{(k)}_0}
    \end{array}
\right]^T
\end{talign}
Where $\boldsymbol{W}^{(k-1)}_0$ is the weight matrix of the feature embedding layer in $(k-1)$th stage and $\boldsymbol{w}_0$ is the expanded proportion for the $k$th task. We then expand the similarity matching layer of the network as: 
\begin{align}
\label{eqILFingerprints}
    \boldsymbol{W_1^{(k)}} = 
\left[
    \begin{array}{c;{2pt/2pt}c}
     \boldsymbol{W_1^{(k-1)}} & \boldsymbol{0}\\ \hdashline[2pt/2pt] 
    \boldsymbol{0} & \boldsymbol{w^{(k)}_1}
    \end{array}
\right]
\end{align}
Where $\boldsymbol{W_1^{(k-1)}}$ is the weight matrix of the similarity matching layer at in the $(k-1)$th stage and $\boldsymbol{w}_1$ is the fingerprints for the $k$th task. The manually inserted zeros on the one hand keep the fingerprints in different stages orthogonal (depicted in Figure~\ref{figCSILSimMatrix}), on the other hand, they enable the feature embedding layer to learn task-specific parameters in different regions (a.k.a. channels). For instance, in Equation \ref{eqILFingerprints}, the newly inserted fingerprints in $\boldsymbol{w^{(k)}_1}$ only make use of embedded features from $\boldsymbol{w^{(k)}_0}$ in Equation \ref{eqILFeatureEmbed}. 

We only train the network with data from the $k$th stage, we use Knowledge Distillation (KD \cite{phuong2019towards}) and Elastic Weight Consolidation (EWC \cite{kirkpatrick2017overcoming}) to prevent the model from forgetting. Therefore, the loss function is defined as:
\begin{align}
\label{eqILLoss}
    \notag L(\boldsymbol{\Theta_{k-1}}, \boldsymbol{\theta_{k}},\boldsymbol{G_m},X_k) = L_{CE} + L_{D} + L_{EWC}
\end{align}
Where $\boldsymbol{\Theta_{k-1}}$ denotes the models' weight at the $(k-1)$th stage. And $\boldsymbol{\theta_{k}} = \{\boldsymbol{w^{(k)}_0}, \boldsymbol{w^{(k)}_1}\}$ denotes the extended weights for the $k$th stage. $\boldsymbol{G_m}$ is a mask matrix, in which the value of each element can only be zero or one. These elements are one-to-one bound to the parameters of a neural network to control which parameter is locked or unlocked. $X_k$ is the training data of the $k$th stage. $L_{CE}$ is the cross entropy loss. $L_{D}$ is the Knowledge Distillation loss:
\begin{align}
    L_{D} = 	\| \boldsymbol{R_{k-1}(X_k)}-\boldsymbol{R_k(X_k)} 	\|
\end{align}
Where $\boldsymbol{R_{k-1}(X_k)}$ is the response of $(k-1)$th model on $X_k$ and $\boldsymbol{R_{k}(X_k)}$ is the response of the $k$th model. $\boldsymbol{F(\cdot)}$ denotes the output of the similarity matching layer ($L_2$). Knowledge Distillation aims to penalize DNNs' behavior from changing drastically.

$L_{EWC}$ in Equation \ref{eqILLoss} represents the Elastic Weight Consolidation (EWC) loss. In EWC, Fisher Information Loss is used to measure the importance of existing parameters, we define EWC Loss for incremental learning as:
\begin{align}
\label{eqContinualLearnLoss}
    L_{EWC}(\boldsymbol{\Theta_k})&=\dfrac{1}{2} \sum _{i}[\boldsymbol{F_{k-1}}\cdot(\boldsymbol{\Theta_k}-\boldsymbol{\Theta_{k-1}})^2]
\end{align}
Where $\boldsymbol{F_{k-1}}$ denotes the Fisher Information (FI) matrix with respect to the $(k-1)$th task. Intuitively, this loss function penalizes the change of critical parameters. The matrix can be estimated as:
\begin{align}
    &\boldsymbol{F_{\boldsymbol{k-1}}}=\left[\dfrac{\partial \log P({X}_{k-1}|\boldsymbol{\Theta_{k-1}})}{\partial \boldsymbol{\Theta_{k-1}}}\right]^2\notag\\
    &P({X}_{k-1}|\boldsymbol{\Theta_{k-1}})\approx\overline{\boldsymbol{Y}_{Softmax}}({X}_{k-1}|\boldsymbol{\Theta_{k-1}})
\end{align}
Where $\overline{\boldsymbol{Y}_{Softmax}}({X}_{k-1}|\boldsymbol{\Theta_{k-1}})$ denotes the averaged outputs of Softmax layer on validation set ${X}_{k-1}$ given parameter set $\boldsymbol{\Omega}$, it approximates the posterior probability $P({X}_{k-1}|\boldsymbol{\Theta_{k-1}})$. $\boldsymbol{F_{\boldsymbol{k-1}}}$ denotes the Fisher information matrix. 

To exemplify the concept, in Figure~\ref{figCSILSimMatrix}, we compare the fingerprints' cosine similarity matrix after regular training and CSIL using the same dataset and scheme specified in Section~\ref{sectConflictsIL}. In this experiment, the convolution layers, the channels for the old task in the feature embedding layer, and the manually supplemented zeros in the similarity matching layer are locked. As a comparison, the DoC of fingerprints is much less apparent compared to Figure~\ref{figCosineSimAftIL}. A more systematic comparison is provided in Section~\ref{sectEED}. 


\section{Performance Evaluation}
In this section, we will evaluate the performance of CSIL algorithm and compare it with the state-of-art.

\label{sectEED}
\subsection{Evaluation dataset}

We use real-world ADS-B signals to verify IL methods for wireless device identification. ADS-B signals are transmitted by commercial aircraft to periodically broadcast their enroute information to Air Traffic Control (ATC) Centers in plain text. These signals are easy to receive and decode but are subject to identity spoofing attacks. We configure our SDR receiver (USRP B210) with a sample rate of 8MHz at 1090MHz, and for each piece of intercepted message, we use the first 1024 complex samples. This dataset is publicly available at \cite{1bxc-ke87-21}. We first decode the ADS-B signals using a modified version of \textit{Gr-ADS-B} in \cite{gt9v-kz32-20} to extract the payloads, then the aircraft's identity codes are used as labels for the truncated messages' signals. We filter out the wireless transponders with less than 500 samples and use the top 100 most frequently seen transponders to construct the dataset. As in Section~\ref{sectZbDNN}, we extract the pseudonoise supplemented with the frequency domain information, we convert each truncated message signal record into a 32 by 32 by 3 tensor. Finally, we got 100 wireless transponders. We use 60\% of the dataset for training and the remaining 40\% of the dataset for validation.

\subsection{Performance comparison}
In this subsection, we compare the CSIL algorithm with other incremental learning algorithms that do not require historical data. The configurations of the selected methods are as follows:
\begin{itemize}
    \item \textbf{Channel Separation Enabled Incremental Learning (CSIL): }We lock the convolution layers and channels in the feature embedding layer which are used by old tasks. We train the new task-specific channels and fingerprints of devices.
    \item \textbf{Learning without Forgetting (LwF): }We lock the convolution layers and the feature embedding layer, we use LwF to train the similarity matching layer.
    \item \textbf{Elastic Weight Consolidation: (EWC)}We lock the convolution layer and feature embedding layer, we train the whole similarity matching layer. The EWC algorithm can adjust old and new fingerprints simultaneously.
    \item \textbf{Finetuning: }We lock the convolution layer, the feature embedding layer, and the old fingerprints, we train the similarity matching layer on new fingerprints.
\end{itemize}
In these configurations, we set the initial learning rate to be 0.01, momentum to 0.9, and $L2$ regularization factor to be 0.01. Stochastic Gradient Descent is selected. We divide the data tensors from 100 wireless devices into 5 batches. We first train the selected DNN model with 20 randomly selected devices and then incrementally train the model with other data batches. During incremental training, the batch size is set to 64 and the models are trained for 10 epoches. 

\begin{figure}[]
\centering  
\subfloat[]
{%
    \includegraphics[width=0.8\linewidth]{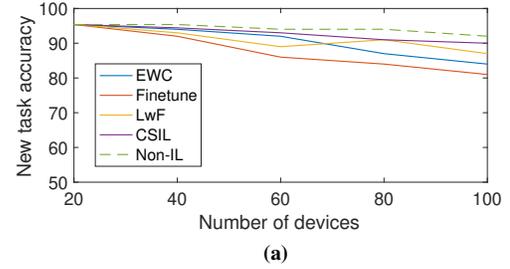}
    \label{figILACCNew}
}\\
\subfloat[]
{%
    \includegraphics[width=0.8\linewidth]{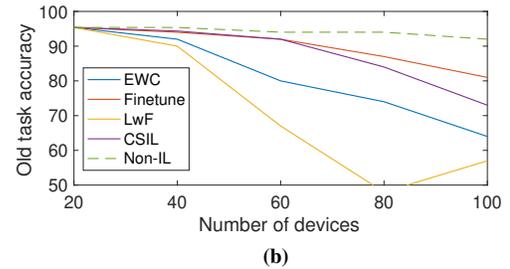}
    \label{figILACCOld}
}\\
\subfloat[]
{%
    \includegraphics[width=0.8\linewidth]{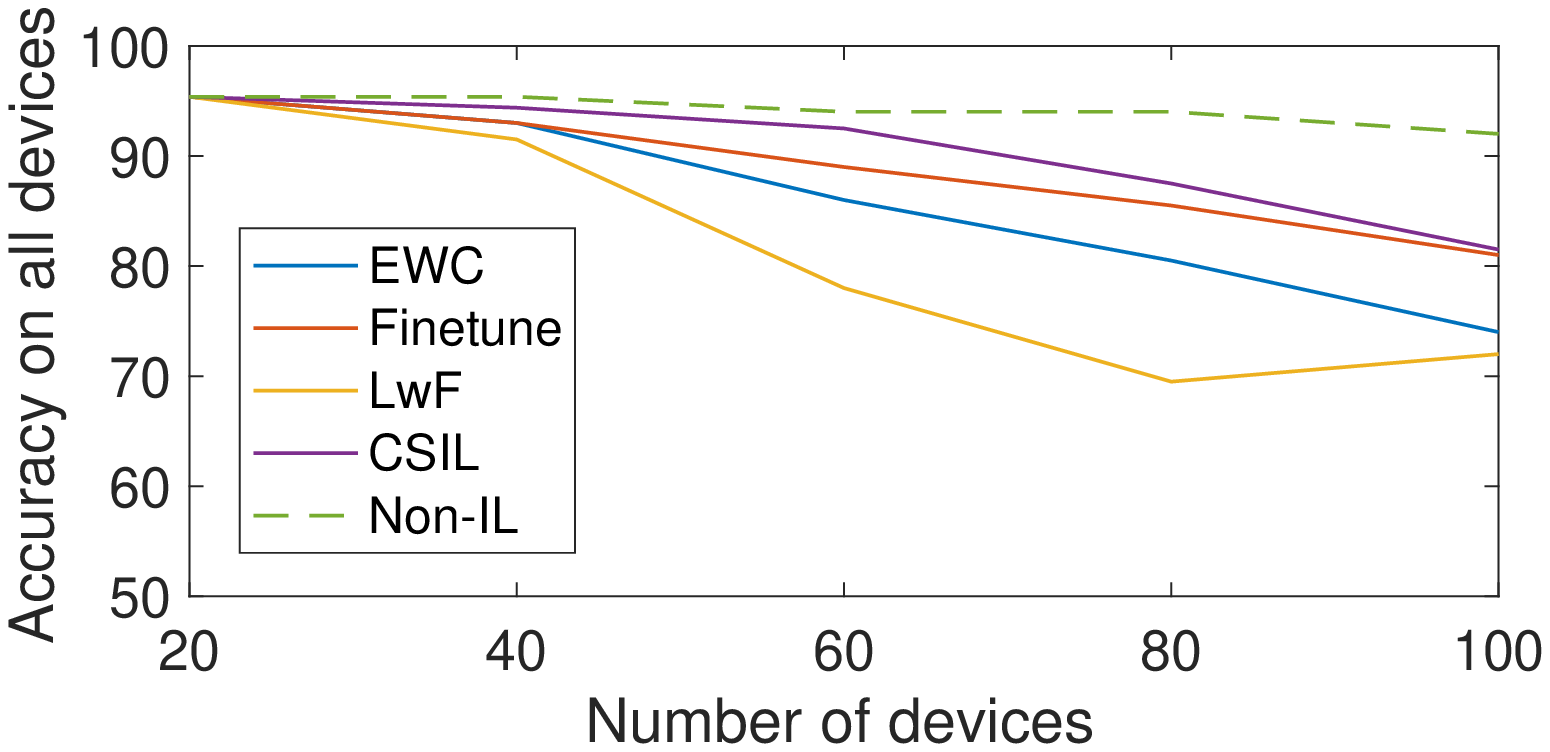}
    \label{figILACCAvg}
}
\caption{Comparison of incremental learning strategies for wireless device identification}
\label{figILScenarios}
\end{figure}

We compare their resulting models' performance on old and incrementally learned new devices as in Figure~\ref{figILScenarios}. Since no historical data is available during incremental learning, forgetting of old tasks are unavoidable. From Figure~\ref{figILACCNew}, the performances of all selected IL algorithms in recognizing new devices are close to the optimal non-IL scheme, in which the proposed CSIL yields the highest accuracy after IL while finetuning with locked old fingerprints shows the worst result. Comparably, in Figure~\ref{figILACCOld}, in preventing forgetting, CSIL's performance is not as good as finetuning with locked weights after learning more than 60 wireless devices (classes). Finetunning with locked weights prevents DNN models from forgetting but with a side effect that prevents the network from learning new devices. The overall performance is given in Figure~\ref{figILACCAvg}, our proposed algorithm CSIL yields the best performance on both old and new tasks.

\begin{figure}[]
    \centering
    \includegraphics[width=0.8\linewidth]{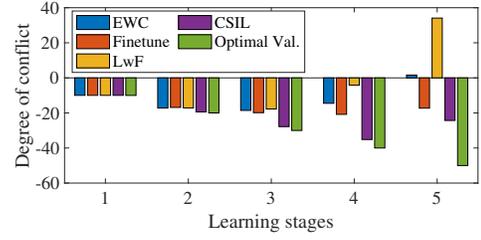}
    \caption{\textcolor{black}{Comparison of Degree of Conflict among IL algorithms}}
    \label{figCompareDoC}
\end{figure}

\textcolor{black}{A comparison of the metric, the Degree of Conflict (DoC), of all devices' fingerprints during incremental learning, is given in Figure~\ref{figCompareDoC}. The propose method, CSIL, yields the lowest DoC. Please be noted that the models' DoC values are still lower than the optimal values (please refer to Equation \ref{eqSumDevResult}) after incremental learning. }

\subsection{Ablation Analysis}
We compare the averaged stage loss of the CSIL considering three factors: a) the Fisher loss. b) The Knowledge Distillation loss. c) The effect of channel separation. The results of ablation analysis are given in Table~\ref{figAblationCSIL}. Apparently, the integral method combining channel separation, EWC, and Knowledge Distillation provides the best performance. 

\begin{table}[b]
\centering
\caption{Ablation analysis of CSIL. All metrics are in percentage.}
\label{figAblationCSIL}
\resizebox{0.5\textwidth}{!}
{%
\begin{threeparttable}
\begin{tabular}{@{}ccllll@{}}
\toprule
Approaches & Initial Acc.\tnote{1} & \multicolumn{1}{c}{\begin{tabular}[c]{@{}c@{}}Acc. with all\\ 100 devices \tnote{2}\end{tabular}} & \multicolumn{1}{c}{\begin{tabular}[c]{@{}c@{}}New acc. at\\ the last stage\end{tabular}} & \multicolumn{1}{c}{\begin{tabular}[c]{@{}c@{}}Old acc. at\\ the last stage\end{tabular}} & \multicolumn{1}{c}{\begin{tabular}[c]{@{}c@{}}Forget / stage \tnote{3} \end{tabular}} \\ \midrule
CS + EWC + KD & 95.2 & \textbf{83.5} & \textbf{90} & \textbf{73} & \textbf{4.5} \\
\sout{CS} + EWC + KD & 95.2 & 75.3 & 82.4 & 66.3 & 5.78 \\
CS + \sout{EWC} + KD & 95.2 & 70.5 & 91 & 50 & 9 \\
CS + EWC + \sout{KD}& 95.2 & 70.5 & 91 & 50.2 & 9 \\

\bottomrule
\end{tabular}%
\begin{tablenotes}

\item[1] \textcolor{black}{Identification accuracy on the first 20 devices, at this stage the network is trained from scratch.}
\item[2] \textcolor{black}{Overall accuracy (100 devices) after the last stage of incremental learning.}
\item[3] \textcolor{black}{Averaged decrease of accuracy on all trained devices after each incremental learning stage.
}
\end{tablenotes}
\end{threeparttable}

}
\end{table}
Notably, without channel separation, the combination of elastic weight consolidation and knowledge distillation can also prevent the network from forgetting. However, such a combination also prevents the network from learning new tasks. Therefore, elastic weight consolidation and knowledge distillation jointly prevent the network from forgetting old devices when training on new tasks, meanwhile, the channel separation mechanism prevents the conflict of class-specific fingerprints. 

\begin{figure}[]
\centering  
\subfloat[]
{%
    \includegraphics[width=0.8\linewidth]{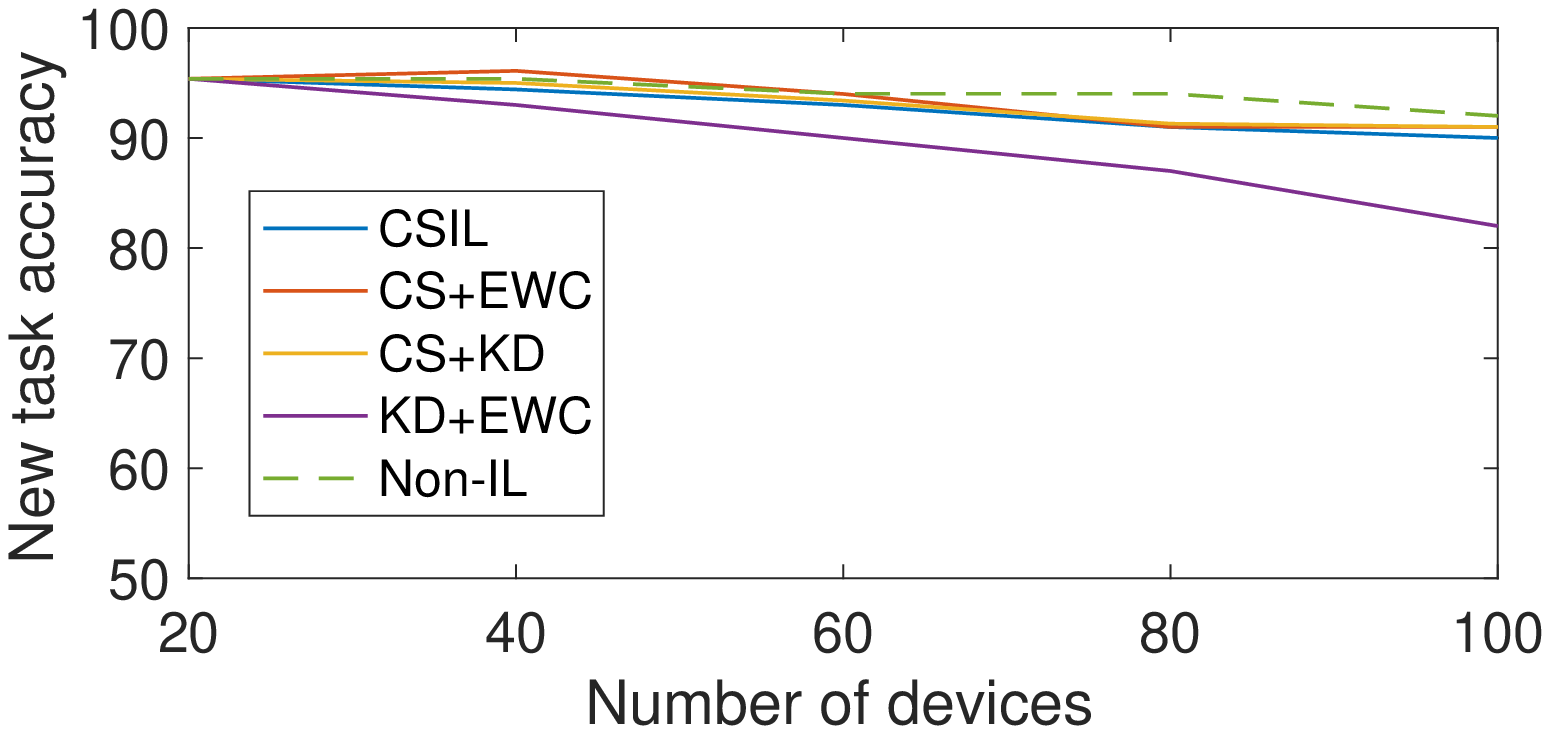}
    \label{figILAblaNew}
}\\
\subfloat[]
{%
    \includegraphics[width=0.8\linewidth]{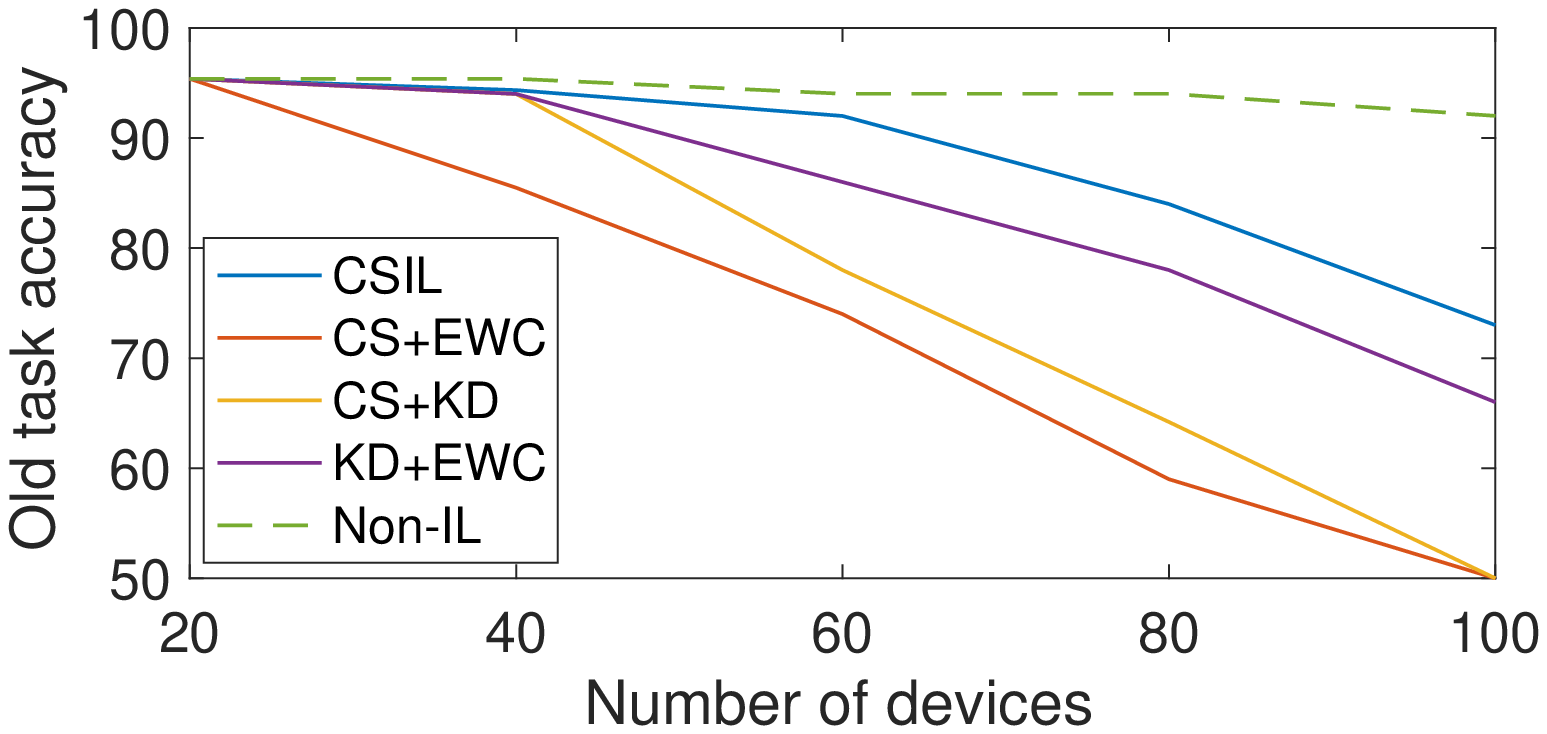}
    \label{figILAblaOld}
}\\
\subfloat[]
{%
    \includegraphics[width=0.8\linewidth]{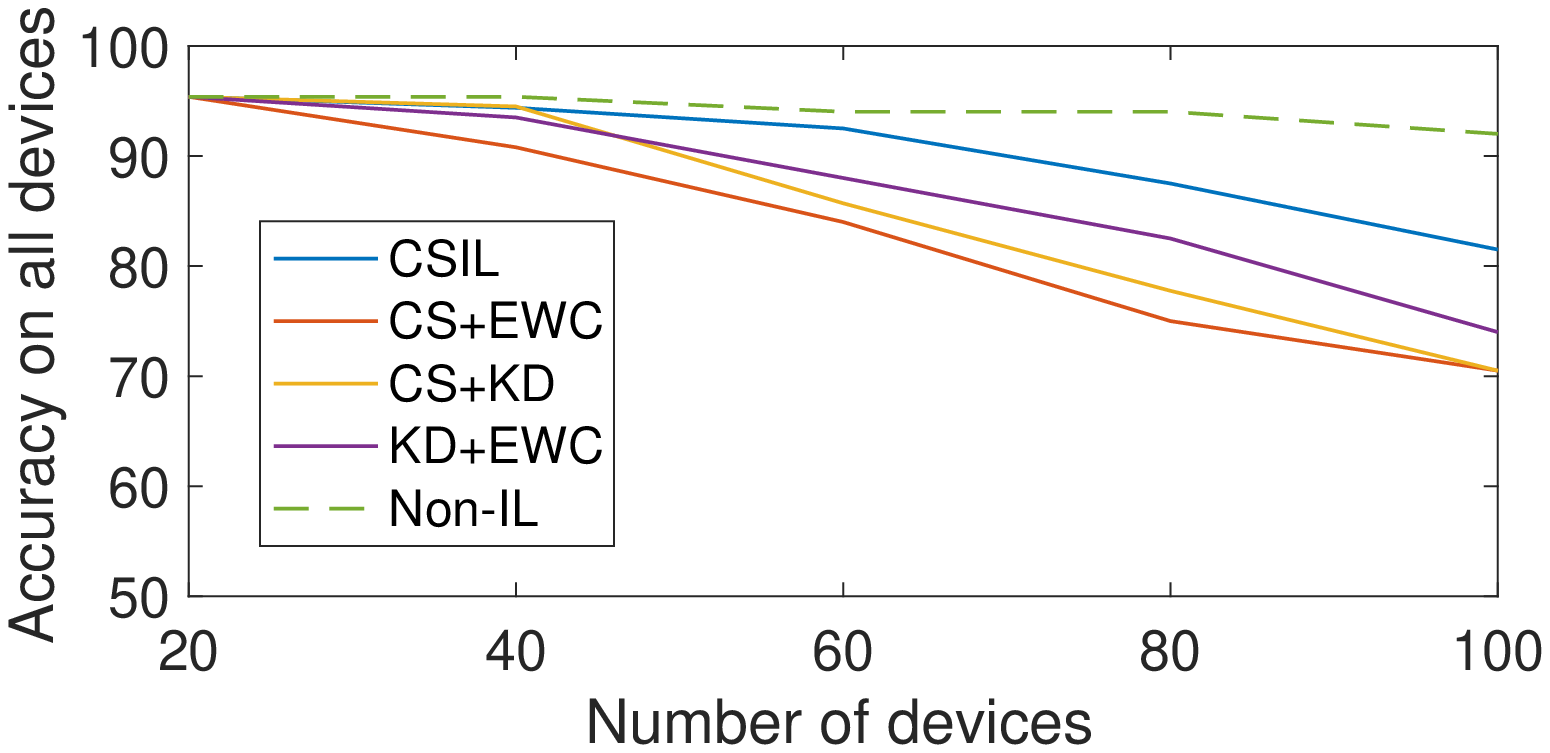}
    \label{figILAblaAvg}
}
\caption{Comparison influential factors in CSIL during incremental learning}
\label{figAblationAnalysis}
\end{figure}

A more detailed comparison is presented in Figure~\ref{figAblationAnalysis}. In Figure~\ref{figILAblaNew}, if the channel separation mechanism is not available, the DNN model will not perform well in learning new devices (classes), as analyzed in Remark~\ref{rmAdjustOld}, the incrementally inserted fingerprints of new devices can conflict with the existing ones, causing the performance degradation. In Figure~\ref{figILAblaOld}, the integral solution, CSIL, yields the highest accuracy in terms of memorizing old devices. Interestingly, the integral of knowledge distillation and elastic weight consolidation ranks the second place in memorizing old devices while showing the worst performance for learning new ones. Therefore, the CSIL provides the best balanced performance between learning and forgetting.

\section{Conclusion}
\label{sectCC}
\textcolor{black}{
In this paper, we propose a novel incremental learning strategy, the Channel Separation Enabled Incremental Learning (CSIL), for wireless non-cryptographic device identification in IoT. Different from existing works, we focus on analyzing the catastrophic forgetting from a new and more thorough perspective, the conflict of device-specific fingerprints. We also propose a novel incremental learning algorithm without using historical data. }

\textcolor{black}{
Our contributions are as follows: Firstly, we provide a new metric, Degree of Conflict (DoC), to measure the degree of topological maturity of DNN models and discover that one important cause for performance degradation in IL is the conflict of classes' representative fingerprints, in which the fingerprints of different devices (classes) are with high cosine similarity, thereby causing confusion. Second, we also show that the conventional IL schemes without using historical data, can lead to DNN models with low topological maturity and high DoC. Thirdly, based on the theoretic analysis, we propose a new IL scheme, the CSIL, based on channel separation and topological control of devices' fingerprints at different stages of learning. We evaluation our proposed solution using the raw signal records from more than 100 aircraft's wireless transponders, and the experiments demonstrate that our CSIL strategy provides the best balance between learning new devices incrementally while retaining the memory of old devices. Therefore, we believe the CSIL and the metric for quantifying the topological maturity of DNN models can be generalized to other domains, such as virus detection or medical image classification. In the future, we will focus on how to better regulate the topological space of DNN models.}

\section*{Acknowledgment}

This research was partially supported by the National Science Foundation under Grant No. 1956193.




\bibliographystyle{IEEEtran}

\bibliography{ReviewRef.bib}
%

%
\vspace{-2em}
\begin{IEEEbiography}[{\includegraphics[width=1in,height=1.25in,clip,keepaspectratio]{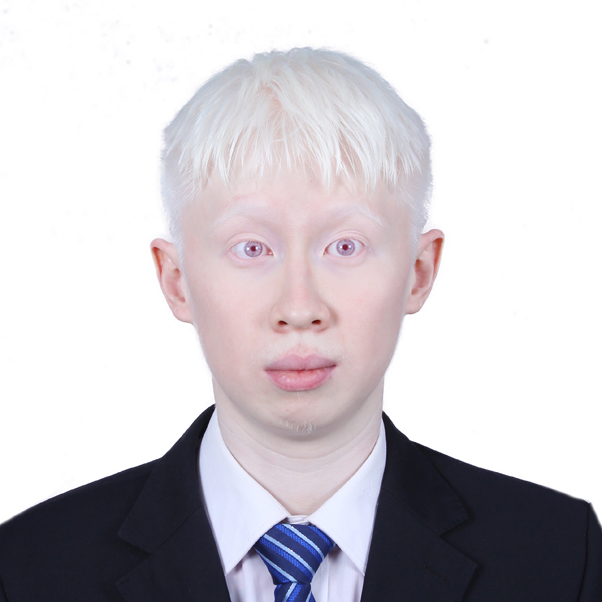}}]{Yongxin Liu}
(LIU11@my.erau.edu) received his first Ph.D. from South China University of Technology in 2018. He is a Ph.D. student in Electrical Engineering and Computer Science at Embry-Riddle Aeronautical University, Daytona Beach, Florida, and a graduate research assistant in the Security and Optimization for Networked Globe Laboratory (SONG Lab, www.SONGLab.us). His major research interests include data mining, wireless networks, the IoT, and unmanned aerial vehicles. 
\end{IEEEbiography}
\vspace{-6em}
\begin{IEEEbiography}[{\includegraphics[width=1in,height=1.25in,clip,keepaspectratio]{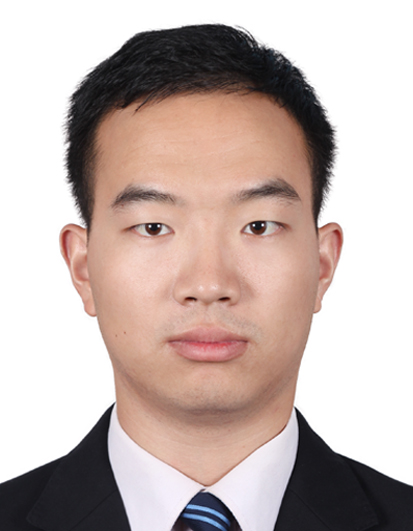}}]{Jian Wang}
(wangj14@my.erau.edu) is a Ph.D. student in the Department of Electrical Engineering and Computer Science, Embry-Riddle Aeronautical University, Daytona Beach, Florida, and a graduate research assistant in the Security and Optimization for Networked Globe Laboratory (SONG Lab, www.SONGLab.us). He received his M.S. from South China Agricultural University in 2017. His research interests include wireless networks, unmanned aerial systems, and machine learning.
\end{IEEEbiography}
\vspace{-5em}
\begin{IEEEbiography}[{\includegraphics[width=1in,height=1.25in,clip,keepaspectratio]{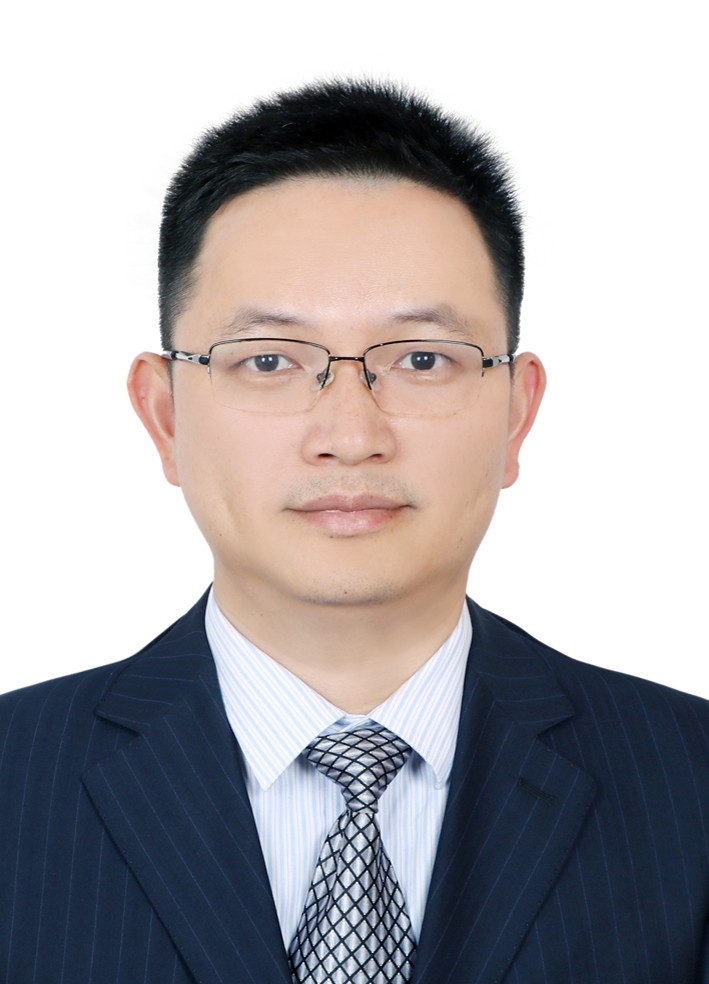}}]{Jianqiang Li}
(lijq@szu.edu.cn) received his B.S. and Ph.D.
degrees from the South China University of
Technology in 2003 and 2008, respectively. He is a Professor with the College of Computer
and Software Engineering, Shenzhen University,
Shenzhen, China. His major
research interests include Internet of Things, robotic,
hybrid systems, and embedded systems.
\end{IEEEbiography}
\vspace{-6em}
\begin{IEEEbiography}[{\includegraphics[width=1in,height=1.1in,clip,keepaspectratio]{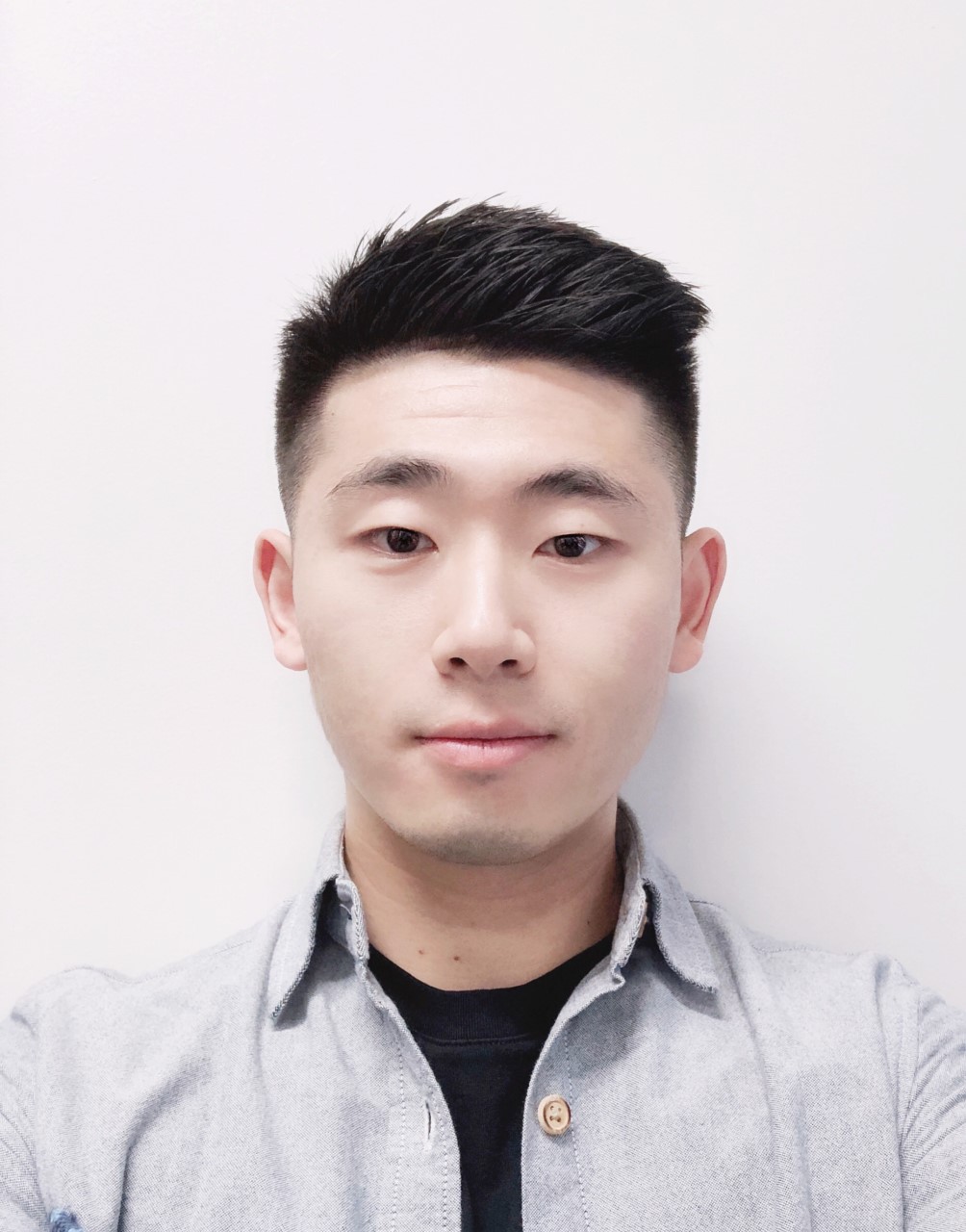}}]{Shuteng Niu}
(shutengn@my.erau.edu) is a Ph.D. student in the Department of Electrical Engineering and Computer Science, Embry-Riddle Aeronautical University (ERAU), Daytona Beach, Florida, and a graduate research assistant in the Security and Optimization for Networked Globe Laboratory (SONG Lab, www.SONGLab.us). He received his M.S. from ERAU in 2018. His research interests include machine learning, data mining, and signal processing.
\end{IEEEbiography}
\vspace{-6em}
\begin{IEEEbiography}[{\includegraphics[width=1in,height=1.25in,clip,keepaspectratio]{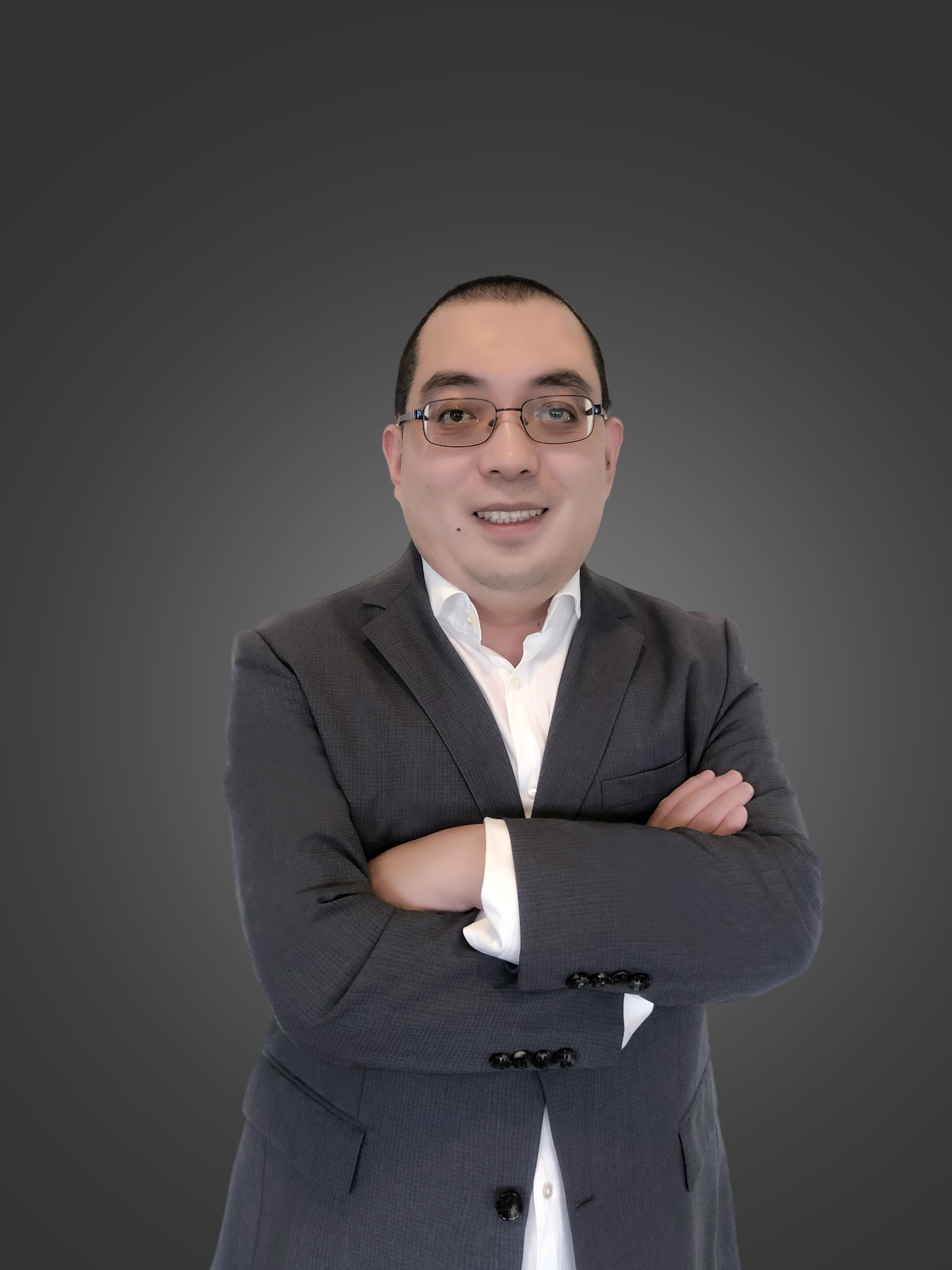}}]{Houbing Song} (M’12–SM’14) received the Ph.D. degree in electrical engineering from the University of Virginia, Charlottesville, VA, in August 2012.

In August 2017, he joined the Department of Electrical Engineering and Computer Science, Embry-Riddle Aeronautical University, Daytona Beach, FL, where he is currently an Assistant Professor and the Director of the Security and Optimization for Networked Globe Laboratory (SONG Lab, www.SONGLab.us). He has served as an Associate Technical Editor for IEEE Communications Magazine (2017-present), an Associate Editor for IEEE Internet of Things Journal (2020-present), IEEE Transactions on Intelligent Transportation Systems (2021-present) and IEEE Journal on Miniaturization for Air and Space Systems (J-MASS) (2020-present). He is the editor of seven books, including Big Data Analytics for Cyber-Physical Systems: Machine Learning for the Internet of Things, Elsevier, 2019, Smart Cities: Foundations, Principles and Applications, Hoboken, NJ: Wiley, 2017, and Industrial Internet of Things, Cham, Switzerland: Springer, 2016.  He is the author of more than 100 articles. His research interests include cyber-physical systems/Internet of things, cybersecurity and privacy, AI/machine learning/big data analytics, and unmanned aircraft systems. His research has been featured by popular news media outlets, including IEEE GlobalSpec's Engineering360, Association for Unmanned Vehicle Systems International (AUVSI), Fox News, USA Today, U.S. News \& World Report, The Washington Times, and New Atlas.

Dr. Song is a senior member of ACM and an ACM Distinguished Speaker. Dr. Song was a recipient of 5 Best Paper Awards (CPSCom-2019, ICII 2019, ICNS 2019, CBDCom 2020, WASA 2020).
\end{IEEEbiography}
\vspace{-5em}




\end{document}